\pdfoutput=1 

\documentclass[runningheads]{llncs}
\usepackage[table,xcdraw,dvipsnames]{xcolor}

\usepackage{graphicx}
\usepackage{tablefootnote}

\usepackage{todonotes}

\usepackage{tabularx}  
\usepackage[pdfstartview=XYZ,
bookmarks=true,
colorlinks=true,
linkcolor=blue,
urlcolor=blue,
citecolor=blue,
pdftex,
bookmarks=true,
linktocpage=true, 
hyperindex=true
]{hyperref}

\usepackage{multirow}
\usepackage{orcidlink}
\usepackage{textcomp}
\usepackage{verbatim}
\usepackage{comment}
\usepackage{threeparttable}
\usepackage{makecell}
\usepackage{xspace}
\usepackage{diagbox}

\newcommand{\ourwork}{LM4HPC\xspace}
\newcommand{\face}{Hugging Face\xspace}
\newcommand{\crawldata}{OMP4Par\xspace}

\begin{document}
\title{\ourwork: Towards Effective Language Model Application in High-Performance Computing}
\titlerunning{LM4HPC: Towards Effective LM Application in HPC}
%
\author{Le Chen\inst{1,2}\orcidlink{0000-0003-3847-4108} \and
Pei-Hung Lin\inst{1}\orcidlink{0000-0003-4977-814X} \and
Tristan Vanderbruggen \inst{1}\orcidlink{0000--0000-0000-0000} \and
Chunhua Liao\inst{1}\orcidlink{0000-0001-6477-0547} \and
Murali Emani \inst{3}\orcidlink{0000--0000-0000-0000} \and
Bronis de Supinski\inst{1}\orcidlink{0000-0002-0339-1006}
}
\authorrunning{Chen et al.}
%
\institute{Lawrence Livermore National Laboratory, Livermore CA 94550, USA \and
Iowa State University, Ames, IA 50010, USA \and 
Argonne National Laboratory, Lemont, IL 60439, USA}
\maketitle              

\setlength{\textfloatsep}{8pt}
\begin{abstract}
In recent years, language models (LMs), such as GPT-4, have been widely used in multiple domains, including natural language processing, visualization, and so on. However, applying them for analyzing and optimizing high-performance computing (HPC) software is still challenging due to the lack of HPC-specific support. In this paper, we design the \ourwork framework to facilitate the research and development of HPC software analyses and optimizations using LMs. Tailored for supporting HPC datasets, AI models, and pipelines, our framework is built on top of a range of components from different levels of the machine learning software stack, with \face-compatible APIs. Using three representative tasks, we evaluated the prototype of our framework. The results show that \ourwork can help users quickly evaluate a set of state-of-the-art models and generate insightful leaderboards.
\keywords{Language model  \and Programming language processing \and High-performance computing}
\end{abstract}
%

\section{Introduction} 

Language models (LMs) are models designed to understand and generate human language. In recent years, large language models (LLMs) 
trained on large amounts of text data have demonstrated stunning capabilities in various natural language processing and visualization tasks. They have also been widely used to process programming languages due to the similarities between natural languages and programming languages.  For example, GPT-4~\cite{bubeck2023sparks} shows early signs of artificial general intelligence. Based on a large language model trained on code~\cite{chen2021evaluating}, GitHub provides an AI assistant for developing software. 

Given the rise of LLMs, it is natural for researchers and developers in the high-performance computing community to start exploiting LMs for addressing various challenges in HPC, including code analysis, code generation, performance optimization, question answering, and so on. However, mainstream frameworks of LMs were originally designed to serve natural language processing. It is difficult for newcomers in HPC to quickly access HPC-specific datasets, models, and pipelines. For example, the current popular \face platform does not include dedicated pipelines for software analyses and optimizations. Another challenge is the entire field is evolving quickly, with new techniques emerging almost weekly, making it challenging for HPC users to keep up with the latest techniques and find relevant ones. Last but not least, there is a lack of standard, reproducible evaluation processes for LMs focusing on HPC-specific tasks. Therefore, it is difficult to have a fair comparison among different models for a given HPC task. 

In this paper, we propose a framework (named \ourwork) designed to serve HPC users as first-class citizens by including internal components and external APIs relevant to HPC-specific tasks. \ourwork's components include models, datasets, pipelines, and so on, while the APIs allow users to interact with these components to finish given HPC tasks. We highlight the contributions of our work as follows:
\begin{itemize}
    \item We design an extensible framework for including and exposing relevant machine learning components to facilitate the adoption of large language models for HPC-specific tasks. 
    \item The framework provides a set of APIs to facilitate essential operations, including code preprocessing, tokenization, integration with new data, and evaluation.
    \item A set of pipelines have been developed to support common HPC tasks, including code similarity analysis, parallelism detection, question answering, and so on.   
    \item We provide HPC-specific datasets such as DRB-ML, \crawldata, and OMPQA to support various HPC pipelines.  
    \item Our work introduces standardized workflows and metrics to enable fair and reproducible evaluation of LLMs for HPC-specific tasks.
    \item Using three representative tasks, we demonstrated how the framework can be used to test a set of language models and generate leaderboards.
\end{itemize}




\section{Background} 
\label{sec-background}
%

Language models (LMs) are machine learning models designed to comprehend and generate human language. They can be used to facilitate natural and intuitive interactions between humans and machines. 
Early generations of LMs, using recurrent neural networks (RNNs),  showed inspiring results for various natural language processing (NLP) tasks.  
A transformative evolution by the Transformer \cite{vaswani2017attention} reveals remarkable potentials of LMs.
Introduced by Vaswani \textit{et al.}, transformer
models utilize the attention mechanism to capture the dependencies between all words in an input sentence, irrespective of their positions. Compared to RNNs, transformers process data in parallel rather than sequentially and significantly improve the efficiency of model training and inference. 
Transformers further enables the inauguration of the large language models (LLMs).
Compared to LMs, LLMs are trained on a vast amount of data and possess parameter counts on the order of billions or more, allowing them to generate more detailed and nuanced responses. Examples of LLMs include OpenAI's GPT-3, GPT-4 and Google's BARD. Nowadays, LLMs have shown remarkable capabilities in NLP tasks like translation, question answering, and text generation.

\vspace{-12pt}
\begin{table}[]
\caption{Language models, associated training data and tasks}
\label{tab-modelExample}
\begin{tabular}{|ccc|cc|c|c|}
\hline
\multicolumn{3}{|c|}{\textbf{Model}} &
  \multicolumn{2}{c|}{\textbf{Training data}} &
  \multirow{2}{*}{\textbf{\begin{tabular}[c]{@{}c@{}}Token\\  Limit\end{tabular}}} &
  \multirow{2}{*}{\textbf{Avail.}} \\ \cline{1-5}
\multicolumn{1}{|c|}{\textbf{Name}} &
  \multicolumn{1}{c|}{\textbf{Released}} &
  \textbf{Size} &
  \multicolumn{1}{c|}{\textbf{Type}} &
  \textbf{Size} &
   &
   \\ \hline
\multicolumn{1}{|c|}{BERT}                     & \multicolumn{1}{c|}{2018/10}       & 340M      & \multicolumn{1}{c|}{Text}  & 3.5B words      & 512    & Weights \\ \hline
\multicolumn{1}{|c|}{CodeBERT} &
  \multicolumn{1}{c|}{2020/11} &
  125M &
  \multicolumn{1}{c|}{Mixed} &
  \begin{tabular}[c]{@{}c@{}}2.1M(bimodal)\\  6.4M (unimodal)\end{tabular} &
  512 &
  Weights \\ \hline
  \multicolumn{1}{|c|}{Megatron}            & \multicolumn{1}{c|}{2021/04}       & 1T      & \multicolumn{1}{c|}{Text}  & 174 GB  & 512    & Weights \\ \hline
\multicolumn{1}{|c|}{GraphCodeBERT}            & \multicolumn{1}{c|}{2021/05}       & 110M      & \multicolumn{1}{c|}{Code}  & 2.3M functions  & 512    & Weights \\ \hline
\multicolumn{1}{|c|}{CodeT5}                   & \multicolumn{1}{c|}{2021/11}       & 770M      & \multicolumn{1}{c|}{Code}  & 8.35M instances & 512    & Weights \\ \hline
\multicolumn{1}{|c|}{GPT-3}                    & \multicolumn{1}{c|}{2022/03/15} & 175B      & \multicolumn{1}{c|}{Mixed} & 500B tokens           & 4096   & Weights \\ \hline
\multicolumn{1}{|c|}{LLaMA}                    & \multicolumn{1}{c|}{2023/02/24}   & 7$\sim$65B     & \multicolumn{1}{c|}{Mixed} & 1.4T tokens      & 4096   & Weights*          \\ \hline
\multicolumn{1}{|c|}{GPT-4}                    & \multicolumn{1}{c|}{2023/03/14}   & 1T        & \multicolumn{1}{c|}{Mixed} & undisclosed     & 8k/32k & API*           \\ \hline
\multicolumn{1}{|c|}{BARD}                     & \multicolumn{1}{c|}{2023/03/21}   & 1.6B      & \multicolumn{1}{c|}{Mixed} & 1.56T words            & 1000   & API           \\ \hline
\multicolumn{1}{|c|}{Cerebras-GPT}             & \multicolumn{1}{c|}{2023/03/28}    & 0.11$\sim$13B & \multicolumn{1}{c|}{Text} & 800 GB           & 2048   & Weights \\ \hline
\multicolumn{1}{|c|}{Dolly 2.0}                & \multicolumn{1}{c|}{2023/04/12}   & 3$\sim$12B     & \multicolumn{1}{c|}{Text}  & 15k instr./resp.             & 2048   & Weights \\ \hline
\multicolumn{1}{|c|}{\begin{tabular}[c]{@{}c@{}}StarCoder\end{tabular}} & \multicolumn{1}{c|}{2023/05/4}   & 15B       & \multicolumn{1}{c|}{Code}  & 1T tokens            & 8192   & Weights \\ \hline
\multicolumn{1}{|c|}{\begin{tabular}[c]{@{}c@{}} StarChat-Alpha\end{tabular}} & \multicolumn{1}{c|}{2023/05/4}   & 16B       & \multicolumn{1}{c|}{Code}  & 31k instr./resp.            & 8192   & Weights \\ \hline
\end{tabular}
\end{table}
\vspace{-15pt}

Table~\ref{tab-modelExample} shows some example language models and their release dates, sizes, training data, input token length limits, and availability. LLaMA~\cite{touvron2023llama}'s weights can be obtained after filling out some form. GPT-4 has a waiting list to use its API. At the time of writing this paper, we have not yet obtained its access.
 

LMs are trained mainly by text data with a focus on NLP.  The sources of the training data mainly come from books, web content, newspapers, scientific articles, and  other text data in various natural languages.    
 Latest LLMs have demonstrated rich skill sets in NLP including text prediction, common sense reasoning, reading comprehension, translation and question answering.

There has been a keen interest in deploying NLP techniques to programming language processing (PLP) tasks, such as code summarization, code generation, and code similarity analysis \cite{chen2022multi,flynn2022finding}. 
Previous studies have demonstrated successful applications of traditional language models to PLP tasks, showing the feasibility of this approach~\cite{devlin2018bert}.
CodeBERT \cite{feng2020codebert}, for example, is a transformer-based model trained with a diverse range of programming languages and can be used for a variety of programming-related tasks. Similarly, CodeT5 \cite{wang2021codet5} is a variant of Google's T5 language model, trained specifically on code datasets to perform advanced programming tasks like code completion, bug detection, and code summarization.
Lately, StarCoder \cite{li2023starcoder}, a 15B parameter model trained with 1 trillion tokens sourced from a large collection of permissively licensed GitHub repositories, is developed to be a Large Language Model mainly for code generation or completion. 
StarChat-Alpha is a GPT-like chat model fine-tuned from StarCoder to act as a helpful coding assistant. 

\vspace{-10pt}
\subsection{LMs for HPC}
With the recent breakthroughs in Generative Pretrained Transformer (GPT) large language models \cite{brown2020language}, it has become increasingly intriguing to explore the application of large language models (LLMs) for HPC tasks.
However, their deployment in the HPC domain is still relatively unexplored. This venture comes with various challenges, including:
\vspace{-4pt}

\begin{enumerate}
    \item Pipelines: Traditional language model frameworks like Hugging Face were designed to support natural language processing or compute vision problems. Expanding LMs to any new domain, including HPC,  requires the addition of new pipelines designed to finish domain-specific tasks.   
    \item Datasets: The HPC domain encompasses an extensive amount of code spanning various fields, including biology and climate modeling. However, preparing this data for machine learning training, such as labeling parallelizable loops in HPC programs for parallelism detection, presents significant challenges. The scarcity of ready-to-use, pre-labeled HPC datasets poses a particular obstacle for training language models, especially large ones, highlighting the need for more shared resources in the community.
    \item Pre-processing: Pre-processing in the context of LMs for HPC typically involves the conversion of source files into a sequence of tokens. However, the direct application of NLP tokenizers to code can be sub-optimal. For instance, an NLP tokenizer might split a variable name into two tokens, a scenario that is not desirable for PLP analysis.  Also, models designed for processing source code may use graph representations, such as abstract syntax trees, to have better performance.
    \item Input size limit: Language models often have limited input token lengths (such as 512 to a few thousand of tokens). HPC tasks often involve processing large-scale software packages with millions of lines of source code. 
    \item Evaluation: There is a pressing need for standardized and reproducible evaluation of different models in the context of various HPC tasks, using metrics suitable for domain-specific requirements. 
\end{enumerate}

\vspace{-14pt}
\section{Approach} 
\label{sec-approach}
To address the challenges discussed in Section~\ref{sec-background}, we introduce \ourwork, a comprehensive framework that encapsulates a suite of machine learning components within user-friendly APIs. This framework is tailored for HPC users, simplifying the implementation process and making the robust capabilities of language models more accessible and user-friendly within the HPC community. The primary goal of \ourwork is to reduce the complexities inherent in employing language models, thus enabling HPC users to leverage their powerful capabilities more effectively and efficiently.

\subsection{\ourwork  design overview}

\begin{figure}[h]
    \centering
    \includegraphics[width=\textwidth]{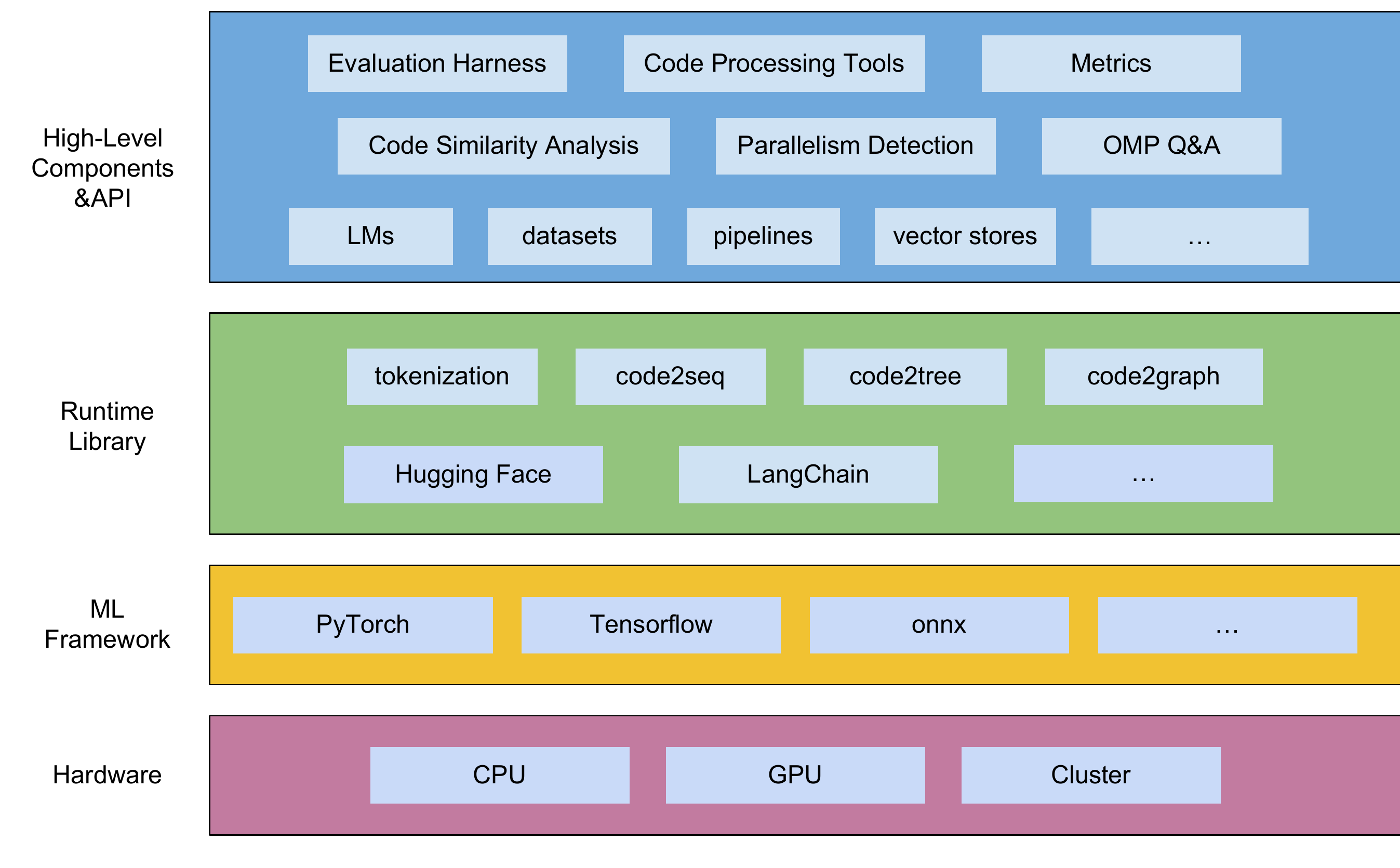}
    \caption{Overview of the \ourwork framework}
    \label{fig:architecture}
\end{figure}

Figure~\ref{fig:architecture} provides the overview of the \ourwork framework. It is built on top of multiple internal machine learning components with \face-compatible APIs. Higher-level components provide concepts and interfaces to users, while middle or lower-level components provide implementation support. 
Table \ref{tab:api_table} shows the example classes and functions in LM4HPC API, including those supporting HPC-specific language models,  tokenizers for programming languages, datasets, inference pipelines, and evaluation.
We elaborate on some essential components in the following subsections. 

\begin{table}[!t]
\centering
\caption{LM4HPC API: example classes and functions. Each class can be imported using ``from lm4hpc import *'' in Python.}
\label{tab:api_table}
\footnotesize 
\setlength{\tabcolsep}{2pt} 
\begin{tabular}{|p{0.13\columnwidth}|p{0.25\columnwidth}|p{0.56\columnwidth}|}
\hline
\textbf{\scriptsize LM4HPC classes} & \textbf{\scriptsize Description} & \textbf{\scriptsize Example API functions} \\
\hline
\multirow{3}{\linewidth}{\scriptsize \texttt{hpcmodel}} &
\multirow{3}{\linewidth}{\raggedright\arraybackslash \scriptsize Fine-tune text-based (HF, OpenAI) and graph-based models, including local private ones, for HPC tasks} &
\raggedright\arraybackslash \scriptsize \texttt{hpcmodel.from\_pretrained(model\_name\_or\_path: Optional[str], *model\_args, **kwargs)} \\ \cline{3-3} 
& & \raggedright\arraybackslash \scriptsize \texttt{hpcmodel.save\_pretrained(model\_name\_or\_path: str, *model\_args, **kwargs)} \\ \cline{3-3} 
& & \raggedright\arraybackslash \scriptsize \texttt{hpcmodel.finetune()} \\
\hline
\multirow{3}{\linewidth}{\scriptsize \texttt{hpctokenizer}} &
\multirow{3}{\linewidth}{\raggedright\arraybackslash \scriptsize APIs to represent code in either tokenized text, trees, or graphs} &
\raggedright\arraybackslash \scriptsize \texttt{hpctokenizer.from\_pretrained(model\_name\_or\_path: Optional[str], *model\_args, **kwargs)} \\ \cline{3-3} 
& & \raggedright\arraybackslash \scriptsize \texttt{hpctokenizer.addtokens(contentsingle\_word=False, strip=False, normalized=True)} \\ \cline{3-3} 
& & \raggedright\arraybackslash \scriptsize \texttt{hpctokenizer.encoding()} \\
\hline
\multirow{4}{\linewidth}{\scriptsize \texttt{hpcdatasets}} &
\multirow{4}{\linewidth}{\raggedright\arraybackslash \scriptsize Load and process HPC datasets} &
\raggedright\arraybackslash \scriptsize \texttt{hpcdatasets.load(path: str, data\_files: Union[str, List, Dict, None], **kwargs)} \\ \cline{3-3} 
& & \raggedright\arraybackslash \scriptsize \texttt{hpcdatasets.split(dataset: hpcdatasets, partition: [float, float, float], **kwargs)} \\ \cline{3-3} 
& & \raggedright\arraybackslash \scriptsize \texttt{hpcdatasets.shuffle(dataset: hpcdatasets, **kwargs)} \\ \cline{3-3} 
& & \raggedright\arraybackslash \scriptsize \texttt{hpcdatasets.sort(dataset: hpcdatasets, **kwargs)} \\
\hline
\scriptsize \texttt{hpcpipeline} &
\scriptsize Pre-built pipelines for common PLP tasks &
\raggedright\arraybackslash \scriptsize \texttt{hpcpipeline(task: str, model\_name\_or\_path: str, *model\_args, **kwargs)} \\
\hline
\multirow{2}{\linewidth}{\scriptsize \texttt{hpceval}} &
\multirow{2}{\linewidth}{\raggedright\arraybackslash \scriptsize Evaluate the results of various models} &
\raggedright\arraybackslash \scriptsize \texttt{hpceval.compute(task: str, models\_name\_or\_path: [[str]], data\_files: Union[str, List, Dict, None], *model\_args, **kwargs)} \\ \cline{3-3} 
& & \raggedright\arraybackslash \scriptsize \texttt{hpceval.plot(shape: str)} \\
\hline
\end{tabular}
\end{table}

\subsection{HPC Tasks and Inference Pipelines}
HPC users are interested in a wide range of tasks related to programming language processing. Table~\ref{tab-HPC-tasks} outlines one way to categorize HPC-specific tasks. The purpose here is not to provide a comprehensive taxonomy of all tasks but a starting point for common tasks we are interested in supporting in our framework. Most tasks are self-explanatory by names and each may have further sub-tasks. For example, clone detection can be viewed as a specialized sub-task under code similarity analysis. 
\begin{table}[htbp]
\centering
\caption{HPC Tasks for Programming Language Processing: Categories and Examples}
\label{tab-HPC-tasks}
\begin{tabular}{l|l|l}
\hline
\textbf{Code Analysis} & \textbf{Code Generation} & \textbf{Others} \\ \hline
Compiler Analysis & Code Completion & Test Case Generation \\
Algorithm Classification & Natural Language-to-Code & Code Search \\
Code Similarity Analysis & Code Translation & Question Answering \\
Documentation Generation & Code Repair & Code Review \\
Parallelism Detection & Code Migration & Decompilation  \\
Defect Detection & Compilation &  IR-to-Source Translation \\ \hline
\end{tabular}
\end{table}

In the context of machine learning, a pipeline represents a sequence of data processing stages to complete a task. Our \ourwork framework extends the pipeline function provided by \face, adapting it for HPC tasks.  
We have developed three inference pipelines: code similarity analysis, parallelism detection, and OpenMP question answering. Code similarity analysis determines the similarity between a pair of code snippets. 
Parallel detection is defined to check if an input code snippet can be parallelized or not using OpenMP.  The OpenMP question answering pipeline is designed to use models to generate answers to OpenMP-related questions. 



Tokenizers are responsible for preprocessing input into an array of numbers as inputs to a model. They are essential components used by pipelines. Most LM tokenizers are primarily designed for NLP tasks. 
For instance, given a function name \texttt{my\_func}, a typical NLP tokenizer like BERT might split it into separate tokens (such as `\texttt{my}', `\texttt{\_}', and `\texttt{func}') while a code-aware tokenizer may treat the function name as a single entity to ensure a more meaningful representation. 

To overcome this, we developed the \ourwork tokenizer, leveraging the treesitter~\cite{treesitter} and programl~\cite{programl} library. Our tokenizer is specifically designed to handle the pre-processing of code data required for a language model. It includes tokenizers such as the ast-tokenizer.
As a result, \ourwork can accommodate models (such as augAST~\cite{chen2023learning-dataset}) that require AST as input in the pipeline. 

\subsection{Datasets}
\label{sec-datasets}
Datasets are crucial for any machine learning application. 
Within the \ourwork framework, we contribute HPC-specific datasets either by converting existing ones into \face-compatible formats or by creating new ones from scratch. 


We have converted three existing datasets to be compatible with \face dataset API: POJ-104~\cite{dataset:poj104}, DRB-ML~\cite{osti_1958879}, and \crawldata~\cite{chen2023learning-dataset}.  
POJ-104 is derived from a pedagogical programming open judge (OJ) system that automatically evaluates the validity of submitted source code for specific problems by executing the code. This dataset is particularly useful for the code similarity task. 
The DRB-ML dataset contains 658 C/C++ OpenMP kernels derived from DataRaceBench~\cite{dataset:dataracebench}. We extended it to have labels indicating if a kernel is parallelizable or not. 
The \crawldata dataset is an open-source benchmark composed of data from three resources: code crawled from GitHub, OpenMP benchmarks such as Nas Parallel Benchmarks~\cite{jin1999openmp} and Rodinia~\cite{che2009rodinia}, and synthetic code. This dataset contains loops with labels indicating whether a loop is parallel and, if parallelizable, the corresponding OpenMP directive associated with the loop. 

Furthermore, we have manually created a new OpenMP question answering dataset called OMPQA in order to probe the capabilities of language models in single-turn interactions with users. Similar to other QA datasets, we include some request-response pairs which are not strictly question-answering pairs. 
The categories and examples of questions in the OMPQA dataset can be found in  Table~\ref{tab-omp-qa}. 

\begin{table}[h]
\centering
\begin{tabular}{|c|c|p{3.7in}|}
\hline
 \textbf{Category}&  \textbf{Count} & \textbf{Example Questions}    \\
\hline
 Basics & 40 &  What is a worksharing construct in OpenMP? \\
\hline
Examples & 20 & Give an example OpenMP C code for computing PI using numerical integration. \\
\hline
Compilers & 24 &  {In what language is LLVM written? \newline How is a parallel region represented in Clang?} \\
\hline
 Benchmarks & 23 & {What are the NAS Parallel benchmarks? \newline Which benchmark assesses data race detection tools?}  \\
\hline 
\end{tabular}
\caption{OMPQA: categories and examples of questions}
\label{tab-omp-qa}
\end{table}

\vspace{-25pt}
\subsection{Integration With New Data}

Language models derive knowledge from training datasets and store this knowledge in internal weights within the model's neural network architecture. However, incorporating new information into a trained model presents a challenge. Traditionally, one might fine-tune pre-trained models with their own data for specific tasks, but this approach requires substantial relevant data and can be resource-intensive. An alternative approach involves integrating new data as context information into a user prompt, but this is constrained by the limited input token lengths of current models.

To address this challenge, \ourwork leverages the LangChain framework~\cite{langchain2023} to easily integrate new data. LangChain aggregates a wide variety of components to build applications using LLMs. Particularly, it provides APIs allowing LLM applications to store large amounts of text in semantic databases called vector stores. 
The way to integrate new data can be done in two steps. First, text data is chunked and embedded with an LLM before being saved into a vector store. 
Later, user prompts are matched with relevant chunks in the vector store using similarity analysis. 
The top-matched chunks are then injected into the original prompts to form a new prompt with relevant context information. By employing this new prompt, language models can generate answers that incorporate new and relevant user data while still staying within the token length limits.

\vspace{-10pt}
\subsection{Evaluation}
An easily accessible harness for evaluating different language models on HPC tasks is crucial. Standard and reproducible results from such evaluations can provide researchers and developers with insightful starting points, helping them select suitable models for their specific needs and identify research opportunities. 

In response to this need, we developed an evaluator API in \ourwork. One challenge we encountered is the lack of standardized metrics for code evaluation. Unlike natural language tasks, where metrics such as BLEU,
ROUGE,
and METEOR
are commonly used, the domain of code lacks such universally accepted measures of quality. We are adding various LLM metrics such as CodeBLEU~\cite{ren2020codebleu} for code output.
Another challenge is that language models may generate different answers for the same input in different inference runs. Evaluation should consider consistent sampling settings (such as temperatures) and control over random seeds to improve reproducibility.



Ultimately, many users are interested in seeing leaderboards that showcase mainstream models competing on common HPC tasks. To satisfy this interest, we create and release a set of test harnesses scripts to enable standard and reproducible evaluation for supported HPC tasks.


\vspace{-10pt}



\section{Preliminary Results} 
\label{sec-evaluation}
In this section, we evaluate the current prototype implementation of LM4HPC through experiments designed to generate leaderboards for three representative tasks: Code Similarity Analysis, Parallelism Detection, and OpenMP Question Answering. \ourwork utilizes LangChain v0.0.174, \face's transformers v4.29.0 and datasets v2.12.0 as our runtime libraries. 
Details of the models and datasets will be discussed in subsequent subsections.

Our experiments were conducted on two machines: 1) a Google Colab VM with a 6-core Xeon processor operating at 2.20GHz, 83.5 GB main memory, 166GB HDD drive, and an NVIDIA A100 GPU with 40 GB memory. 2) a Dell workstation equipped with a dual Intel Xeon 6238 CPU operating at 2.10GHz, 128 GB main memory, 1TB SSD drive, and an NVIDIA Quadro RTX 6000 GPU with 24GB memory.  The majority of our experiments were run on the Google Colab machine to leverage its superior GPU memory. However, we encountered difficulties running Cerebras-GPT on the Colab machine and were compelled to use the Dell workstation with larger CPU memory instead. 
\subsection{Code Similarity Analysis}
The code similarity task is designed to measure the syntactic and/or semantic similarity between two code snippets. Such analysis information can be beneficial in various scenarios such as plagiarism detection, code reuse and refactoring, bug detection and repair, licensing compliance, malware detection, and so on.

\textbf{Preparing Datasets and Ground Truth.}
Two datasets introduced in Section~\ref{sec-datasets}, POJ-104 and DRB-ML, are loaded through \ourwork's datasets API for this experiment.
For each pair of code snippets in the POJ-104 dataset, we assign a binary similarity label based on their functional labels.  A similarity label of 1 signifies that the snippet pair shares the same functional label and we assign a similarity score of 1. Otherwise, the label is 0. 
We have processed the DRB-ML dataset using a similar methodology to generate code pairs and labels. The main difference is that the similarity ground truth for DRB-ML is derived from its own similarity score table~\cite{chen2022early}, providing a precise and reliable similarity measurement between code snippets in the dataset.

\textbf{Inference Experiments and Evaluation.}
We employ \ourwork's code similarity pipeline to test various models. 
The default model for this pipeline is CodeBERT. We additionally select four models from Table \ref{tab-modelExample} for evaluation: GraphCodeBERT, gpt-3.5-turbo, Dolly 2.0 (12B), and Cerebras-GPT (13B). 
We set the maximum token length for the model output to 256. This limits the verbosity of the model and keeps its responses concise. Additionally, we set the temperature parameter to 0 when applicable. For models like Dolly 2.0 that require positive temperature, we set the temperature to be $1\times 10^{-6}$. This setting ensures that the model's responses are consistent and deterministic, minimizing variability and uncertainty in its output. 


Within \ourwork, the approach of processing input code pairs depends on the type of the model employed. 
Models like CodeBert and GraphCodeBert are specifically devised and trained on a variety of programming languages. We directly feed a pair of code snippets to generate a similarity prediction.
On the other hand, large language models like gpt-3.5-turbo, Dolly 2.0, and Cerebras-GPT are evaluated using the following prompt template: ``Code 1: \{...\} Code 2: \{...\}  Determine whether the two code snippets are similar. If the code snippets are similar, output 1; otherwise, output 0.''.



\textbf{Results.}
The Code Similarity Analysis leaderboards generated using the two datasets are shown in Table~\ref{tab:hpctask1}.
Notably, gpt-3.5-turbo demonstrates superior performance.
Two other models, StarChat-Alpha and Dolly 2.0, also exhibit commendable performance. 
Most large language models outperform traditional models (GraphCodeBERT and CodeBERT) that were specifically trained for code analysis. 
However, Cerebras-GPT struggled to comprehend the code and mostly returned arbitrary word tokens, indicating a lack of effective code understanding since it is mostly designed for natural language processing.
\begin{table}[htbp]
\caption{Code Similarity Analysis Leaderboard: POJ-104 (left) and DRB-ML (right)}
\label{tab:hpctask1}
\begin{minipage}{.5\linewidth}
  \begin{tabular}{|c|c|c|c|}
  \hline
  \textbf{Model} & \textbf{Precision}              & \textbf{Recall}                 & \textbf{F1}                     \\ \hline
  \textbf{gpt-3.5-turbo}     & \textbf{78.4}                   & \textbf{74.2}                   & \textbf{76.2}         \\ \hline
  Dolly 2.0 12B  & 61.9                   & 61.3                   & 61.6                   \\ \hline
StarChat-Alpha  & 59.4                      & 56.2                      & 57.8            \\ \hline
GraphCodeBERT   & 52.7                   & 60.3                   & 56.3             \\ \hline
CodeBERT        & 51.5                   & 59.4                   & 55.2          \\ \hline
Cerebras-GPT 13B   & 0                      & 0                      & 0              \\ \hline        
\end{tabular}
\end{minipage}
\hspace{0.1cm}
\begin{minipage}{.5\linewidth}
  \begin{tabular}{|c|c|c|c|}
\hline
\textbf{Model} & \textbf{Precision}              & \textbf{Recall}                 & \textbf{F1}                     \\ \hline
\textbf{gpt-3.5-turbo}     & \textbf{82.4}                   & \textbf{81.3}                   & \textbf{81.8}           \\ \hline
StarChat-Alpha  & 79.6                   & 77.4                   & 78.5                   \\ \hline
Dolly 2.0 12B  & 74.3                      & 73.2                      & 73.7            \\ \hline
GraphCodeBERT   & 79.4                   & 77.9                   & 78.6            \\ \hline
CodeBERT        & 76.9                   & 74.5                   & 75.7          \\ \hline
Cerebras-GPT 13B    & 0                      & 0                      & 0              \\ \hline            
\end{tabular}  
\end{minipage}
\end{table}

\subsection{Parallelism Detection}
\vspace{-5pt}
The parallelism detection task aims to identify parallelism opportunities within a given code snippet. We utilized two datasets, \crawldata  and DRB-ML introduced in Section~\ref{sec-datasets}, for the experiment. 


\textbf{Preparing Datasets and Ground Truth.}
The \crawldata dataset is specifically designed for parallelism detection. 
Its existing labeling scheme allows us to prepare the data for binary classification models. 
Similarly, we prepared DRB-ML dataset with a label indicating whether each code snippet is parallelizable using OpenMP or not.  

It is worth noting that both datasets have undergone source code pre-processing steps, including comment removal and code snippet extraction. These steps are common practice~\cite{chen2023learning-dataset} to ensure that code snippets are small enough to be fed into language models with limited input token sequence sizes. However, the resulting code snippets may lose their context information, such as variable declarations. This is a serious limitation of language models with limited input sizes when applied to process large source files. 




\textbf{Inference Experiments and Results.}
We selected six models to generate parallelism detection leaderboards. Four of them are introduced in Section~\ref{sec-background}.  They take the code snippets in a prompt template: ``As an OpenMP expert, you will analyze the given code snippet to determine if it can be parallelized. Code: \{...\}. Answer yes or no first:''. The other two are augAST~\cite{chen2023learning-dataset} and DeepSCC-based~\cite{harel2023learning}, which are pre-trained models using \crawldata's training dataset. We fed code snippets to these two models to directly obtain predicted labels. 

Table \ref{tab:t2} presents the resulting leaderboards. 
The highest F1 score reaches 93.9, indicating that LMs can be very effective for detecting parallelism. However, the datasets contain small-scale code snippets that are significantly simpler than real HPC codes. 
Again, gpt-3.5-turbo outperforms all other models overall, including specially trained models like augAST and DeepSCC.
AugAST performs better than gpt-3.5-turbo in terms of precision, suggesting it's more effective in predicting a positive class, which, in this case, is parallelizable code. 
Finally, Cerebras-GPT did not perform well in this code analysis task. 


\begin{table}[htbp]
  \centering
  \caption{Parallelism Detection Leaderboards: \crawldata(left) and DRB-ML(right)}
\label{tab:t2}
  \begin{minipage}{.5\linewidth}
    \centering
    \resizebox{6cm}{!}{
\begin{tabular}{|c|c|c|c|c|}
\hline
\textbf{Model} & \textbf{Precision}              & \textbf{Recall}                 & \textbf{F1}                     \\ \hline
\textbf{gpt-3.5-turbo}     & 90.6                   & \textbf{89.3}                  & \textbf{89.9}                   \\ \hline
augAST        & \textbf{92.1}                   & 82.4                   & 87.0                   \\ \hline
DeepSCC   & 82.7                   & 81.4                   & 82.0                   \\ \hline
StarChat-Alpha   & 85.7                   & 68.2                  & 75.9                   \\ \hline
Dolly 2.0 12B      & 64.2                   & 63.7                   & 63.9                      \\ \hline
Cerebras-GPT 13B    & 0                      & 0                      & 0
    \\ \hline
\end{tabular} 
}
  \end{minipage}%
  \hspace{-1.13cm}
  \hspace{0.9cm}
   \begin{minipage}{.5\linewidth}
    \centering
    \resizebox{6cm}{!}
{
\begin{tabular}{|c|c|c|c|c|}
\hline
\textbf{Model} & \textbf{Precision}              & \textbf{Recall}                 & \textbf{F1}                     \\ \hline
\textbf{gpt-3.5-turbo}     & 90.0                   & \textbf{98.9}                  & \textbf{94.2}                   \\ \hline
augAST        & \textbf{91.4}                   & 72.3                   & 80.7                   \\ \hline
DeepSCC   & 80.4                   & 79.5                   & 79.9                   \\ \hline
StarChat-Alpha   & 81.9                   & 20.3                  & 32.5                   \\ \hline
Dolly 2.0 12B      & 40.0                   & 11.2                   & 2.17                      \\ \hline
Cerebras-GPT 13B    & 0                      & 0                      & 0
    \\ \hline
\end{tabular}
}
  \end{minipage}%
\end{table}

\subsection{OpenMP Q\&A}
 In this experiment, we utilized \ourwork to evaluate the capabilities of several language models in answering questions related to OpenMP. This evaluation was conducted using the OMPQA dataset, introduced in Section \ref{sec-datasets}. 

\textbf{Experiment Settings.} 


Each model receives the question in the following prompt template:
``You are an OpenMP expert. Please answer this question. Question: \{question\}''. Two metrics are selected to evaluate the quality of answers: the Bilingual Evaluation Understudy (BLEU) and ROUGE-L metrics. BLEU is a precision-oriented metric measuring the overlap of n-grams between the generated text and a set of reference texts. 
ROUGE-L (Recall-Oriented Understudy for Gisting Evaluation - Longest Common Subsequence) calculates the longest common subsequence (LCS) that appears in a left-to-right sequence in both the system-generated and reference summaries, thus providing a measure of the coherence and fluidity of the generated text.

\textbf{Results.} 
Table \ref{tab:t3-qa} displays the Q\&A leaderboard of several selected models. We additionally include the memory and execution time information. The experiments using gpt-3.5-turbo do not consume any local GPU memory since the model is invoked remotely through OpenAI's API.

Again, gpt-3.5-turbo unsurprisingly outperforms other LLMs, including StarChat-Alpha, Dolly 2.0, and Cerebras-GPT. However, the highest ROUGE-L F1 score of 0.259 indicates that all models have room for improvement in answering OpenMP questions. One reason is that many questions in OMPQA are open-ended and do not necessarily have a single correct answer. Also, the two metrics used do not sufficiently consider semantics.


\begin{table}[]
\caption{Q\&A Leaderboard using the OMPQA dataset. The arrow indicates the performance changes when augmenting external knowledge base by LangChain.}
\label{tab:t3-qa}
\begin{tabular}{|c|c|c|c|c|ccc|}
\hline
 &
   &
   &
   &
   &
  \multicolumn{3}{c|}{\textbf{ROUGE}} \\ \cline{6-8} 
\multirow{-2}{*}{\textbf{Model}} &
  \multirow{-2}{*}{\begin{tabular}[c]{@{}c@{}}\textbf{\scriptsize CPU}\\ \scriptsize  Mem. (GB)\end{tabular}} &
  \multirow{-2}{*}{\begin{tabular}[c]{@{}c@{}}\textbf{\scriptsize GPU} \\ \scriptsize  Mem. (GB)\end{tabular}} &
  \multirow{-2}{*}{\textbf{Time}(s)} &
  \multirow{-2}{*}{\textbf{BLEU}} &
  \multicolumn{1}{c|}{Recall} &
  \multicolumn{1}{c|}{Prescision} &
  F1 \\ \hline
  \begin{tabular}[c]{@{}c@{}}gpt-3.5-turbo\\ + LangChain\end{tabular}
 &
  4.1 &
  0 &
  21.452 &
  \textbf{{0.147}$\uparrow$} &
  \multicolumn{1}{c|}{{0.347$\downarrow$}} &
  \multicolumn{1}{c|}{0.262$\downarrow$} &
  \textbf{0.259$\uparrow$} \\ \hline
gpt-3.5-turbo &
  4.2 &
  0 &
  12.749 &
  0.139 &
  \multicolumn{1}{c|}{\textbf{0.446}} &
  \multicolumn{1}{c|}{0.231} &
  0.257 \\ \hline
StarChat-Alpha &
  6.8 &
  18.9 &
  29.732 &
  0.082 &
  \multicolumn{1}{c|}{0.322} &
  \multicolumn{1}{c|}{0.149} &
  0.172 \\ \hline
  \begin{tabular}[c]{@{}c@{}}Dolly 2.0 12B\\ + LangChain\end{tabular}
  &
  27.4 &
  39.8 &
  7.217 &
  0.084$\uparrow$&
  \multicolumn{1}{c|}{0.228$\uparrow$} &
  \multicolumn{1}{c|}{0.232$\downarrow$} &
  0.182$\uparrow$ \\ \hline
Dolly 2.0 12B &
  27.1 &
  39.2 &
  8.147 &
  0.06 &
  \multicolumn{1}{c|}{0.208} &
  \multicolumn{1}{c|}{\textbf{0.312}} &
  0.148 \\ \hline
Cerebras-GPT 13B &
  52.6 &
  11.7 &
  590.763 &
  0.071 &
  \multicolumn{1}{c|}{0.319} &
  \multicolumn{1}{c|}{0.089} &
  0.112 \\ \hline
\end{tabular}
\end{table}

To enhance the capacity of large language models (LLMs) in accurately responding to OpenMP queries, we integrate the official OpenMP documentation into our process. 
We employ LangChain, a mechanism designed to efficiently store and retrieve language model embeddings, enabling us to accommodate large volumes of new data. To assess the efficacy of using LangChain to incorporate additional user data, we leverage its API to create a vector store. This vector store holds embeddings of text chunks derived from the OpenMP API Specification v5.2 (669 pages) and the OpenMP Application Programming Interface Examples v5.2.1 (575 pages). We then select two LangChain-supported models, GPT-3.5 and Dolly 2.0, to utilize the vector store as an additional resource for answering queries, thereby demonstrating the practical utility of our approach.
The results indicate slight improvements in both the BLEU and ROUGE-L F1 scores, increasing from 0.139 to 0.147 and from 0.257 to 0.259, respectively. However, there are mixed results for recall and precision metrics. gpt-3.5-turbo has a better recall, 0.446, compared to 0.347 of the Langchain approach.  

Further, we examine the effectiveness of the LangChain approach across different question categories. When addressing `Basic' questions, the BLEU scores rise by 20.7\% and 9.8\% for gpt-3.5-turbo and Dolly 2.0, respectively. Additionally, we assess the LangChain performance using the CodeBLEU metric\cite{ren2020codebleu} for the `Examples' category, observing a score increase of 6.1\% and 12.2\% for gpt-3.5-turbo and Dolly 2.0, respectively. These observations indicate that augmenting LLMs with documentation via LangChain improves performance for `Basic' and `Examples' categories. However, for `Compilers' and `Benchmarks' categories, the performance of gpt-3.5-turbo and Dolly 2.0 diminishes when utilizing LangChain, recording an average BLEU score drop of 8.0\% and 7.9\%, respectively. This drop is likely because our documentation does not include information relevant to compiler and benchmark topics.

We also manually investigated the answers generated by the models. Overall, StarChat-Alpha delivers competitive results compared to GPT-3.5. It seems to be a good choice for people who want to use open-source language models based on our experiments. Research has indicated that GPT-4 surpasses GPT-3.5 in a variety of domains. However, as of now, API accessibility for GPT-4 has not been made publicly available. We plan to assess GPT-4's performance as soon as it becomes accessible and incorporate it into our framework if it benefits HPC tasks.

\section{Related Work} 
PyTorch and TensorFlow are the most popular frameworks, backed by Meta AI and Google, respectively.
Both frameworks are similar in many respects, including 1) providing low-level APIs for development, 2) supporting a rich collection of libraries, and 3) maintaining dedicated hubs - PyTorch Hub and TensorFlow Hub - for providing pre-trained ML models. 
\face is a large open-source community that builds tools to enable users to build, train, and deploy machine learning models based on open-source code and technologies.
\face is best known for its \texttt{Transformers} library, which exposes a collection of Python APIs to leverage state-of-the-art deep learning architectures for NLP tasks. 
With the goal to simplify end-to-end NLP tasks, \face \texttt{Transformers} offers a pipeline that performs all pre- and post-processing steps on the given input text data. The overall process of the model inference is encapsulated within these pipelines. With the pipeline, users only need to provide the input texts and the model for the task.  The remaining connections among a model and required pre- and post-processing steps are hidden within the pipeline implementation.

There were various research works and developments to improve the ML ecosystem to be Findable, Accessible, Interoperable, and Reproducible (FAIR).  These existing frameworks aim to make the models, datasets, or both FAIR.  Among these frameworks, HPCFAIR~\cite{verma2021hpcfair} focuses on providing support for model interoperability, search capabilities for datasets and models, and seamless integration into HPC workflows.  The work in \cite{yu2022seamless} extended this work to include support for interoperability across different framework implementations using ONNX and provision to retrain a model with transfer learning.
However, HPCFAIR framework relies on users to handle data pre- and post-processing. In comparison, LM4HPC is equipped to manage data processing within the pipeline design and generate leaderboards for supported HPC tasks. 


General LLMs are trained with data covering general knowledge and information that is usually collected from public domains. 
Domain-specific datasets can be collected for the training of a specialized model or for the fine-tuning of a general-purpose model.
MedQA\cite{jin2020disease} is an example of domain-specific datasets collecting question-answer pairs and textbooks  from professional medical board exams.
ExeBench~\cite{exebench}, another domain-specific dataset for tasks in compilation and software engineering, contains millions of runnable and representative C functions collected from GitHub.
In addition to collecting existing data, ML research has started to automate dataset creation with assistance from the LLMs.  
The developers of LaMini-LM~\cite{wu2023lamini} develop a large set of 2.58M instruction and response pairs based on both existing and newly-generated instructions.  A handful of seed examples from the existing LLM prompts and 2.2M categories from Wikipedia from existing  are submitted to the \texttt{gpt-3.5-turbo} to generate relevant instructions.  Similarly, the responses for the generated instructions are also generated by the \texttt{gpt-3.5-turbo}.


\label{sec-related-work}

\section{Conclusion} 
In this paper, we presented our efforts to facilitate the application of language models for tasks specific to High-Performance Computing. We have developed the LM4HPC framework to encompass and expose relevant machine learning components via corresponding APIs.
Our experimental findings suggest that GPT-3 performs competitively, despite not being specifically designed for HPC tasks. However, there is significant room for improvement in answering OpenMP questions. Furthermore, the input size limitation of language models adds complexity to certain tasks, such as parallelism detection. Finally, an obstacle to advancing the application of language models for HPC tasks is the absence of HPC-specific training and evaluation datasets.

Looking ahead, our future work will explore automated approaches to generating HPC-specific datasets. We intend to enhance LM4HPC's capabilities to support the fine-tuning of models for HPC-related tasks, including those related to the Message Passing Interface (MPI), and to provide performance analysis and optimization suggestions.

\label{sec-conclusion}

\section*{Acknowledgement}
Prepared by LLNL under Contract DE-AC52-07NA27344 (LLNL-CONF-849438) and supported by the U.S. Department of Energy, Office of Science, Advanced Scientific Computing Research.


\bibliographystyle{splncs04}
\bibliography{main.bib}

\end{document}